\documentclass[runningheads]{llncs}
\usepackage{graphicx}
\usepackage{xcolor} 
\usepackage[framemethod=tikz]{mdframed}

\usepackage[pagebackref=false,breaklinks=true,colorlinks,bookmarks=false]{hyperref}
\usepackage{amssymb}

\newcommand*\samethanks[1][\value{footnote}]{\footnotemark[#1]}

\begin{document}
\title{Latent-Graph Learning for Disease Prediction}


%\author{First Author\inst{1}\orcidID{0000-1111-2222-3333} \and
%Second Author\inst{2,3}\orcidID{1111-2222-3333-4444} \and
%Third Author\inst{3}\orcidID{2222--3333-4444-5555}}

\authorrunning{Cosmo et al.}
\author{Luca Cosmo\thanks{Equal contribution}\inst{1,2} \and Anees Kazi\samethanks{}\inst{3} \and
Seyed-Ahmad Ahmadi\inst{4} \and Nassir Navab\thanks{Shared last authorship} \inst{3,5}\and Michael Bronstein\samethanks{}\inst{1,6,7}}
% \index{Cosmo, Luca}
% \index{Kazi, Anees}
% \index{Ahmadi, Seyed-Ahmad}
% \index{Navab, Nassir}
% \index{Bronstein, Michael}

\institute{University of Lugano, Switzerland\\
\and
Sapienza University of Rome, Italy\\
\and
Computer Aided Medical Procedures (CAMP), Technical University of Munich, Germany\\
\and
German Center for Vertigo and Balance Disorders, Ludwig Maximilians Universit\"at M\"unchen, Germany\\
\and
Whiting School of Engineering, Johns Hopkins University, Baltimore, USA\\
\and
Imperial College London, UK\\
\and 
Twitter, UK\\
\email{luca.cosmo@usi.ch}
}

\maketitle              % typeset the header of the contribution
\begin{abstract}
Recently, Graph Convolutional Networks (GCNs) have proven to be a powerful machine learning tool for Computer Aided Diagnosis (CADx) and disease prediction. A key component in these models is to build a population graph, where the graph adjacency matrix represents pair-wise patient similarities. Until now, the similarity metrics have been defined manually, usually based on meta-features like demographics or clinical scores. The definition of the metric, however, needs careful tuning, as GCNs are very sensitive to the graph structure. 
In this paper, we demonstrate for the first time in the CADx domain that it is possible to learn a single, optimal graph towards the GCN's downstream task of disease classification. To this end, we propose a novel, end-to-end trainable graph learning architecture for dynamic and localized graph pruning. Unlike commonly employed spectral GCN approaches, our GCN is spatial and inductive, and can thus infer previously unseen patients as well. We demonstrate significant classification improvements with our learned graph on two CADx problems in medicine. We further explain and visualize this result using an artificial dataset, underlining the importance of graph learning for more accurate and robust inference with GCNs in medical applications.
\end{abstract}
\section{Introduction}
There is a growing body of literature that recognises the potential of geometric deep learning \cite{bronstein2017geometric} and graph convolutional networks (GCNs) in healthcare. GCNs have been applied to various problems already, including protein interaction prediction \cite{gainza2019deciphering}, metric learning on brain connectomes \cite{ktena2018metric} or representation learning for medical images \cite{burwinkel2019cnngat}.
In this work, we focus on Computer Aided Diagnosis (CADx), i.e. disease prediction from multimodal patient data using GCNs in conjunction with population models \cite{parisot2018disease,vivar2018multi,kazi2019self,kazi2019graph}.
The population model is realized in form of a graph, where vertices represent patients, edges represent connections between patients, and edge weights represent the patient similarity according to a defined metric \cite{parisot2018disease}. Employing graph models for CADx is motivated by the success of GCNs in social network analyses and recommender systems \cite{bronstein2017geometric}. The graph provides neighborhood information between patients, and graph signal processing \cite{ortega2018gsp} is used to aggregate patient features over local neighborhoods, similar to localized filters in regular convolutional neural networks (CNN). Graph deep learning is then tasked with learning a set of filters with optimal weights towards the downstream task of CADx classification.
Kipf and Welling  \cite{kipf2016semi} proposed semi-supervised node classification using spectral GCNs. This approach was first adapted to CADx in medicine by Parisot et al. \cite{parisot2018disease}, who proposed to %build a single populating
compute patient similarities from a set of meta-features such as age and sex. Importantly, the definition of the similarity metric between patients determines the graph adjacency matrix.
As such, both feature selection and similarity metric definition need to be carefully tuned to avoid placing a meaningless graph structure at the core of the GCN.
To alleviate this, several works have proposed using multiple graphs \cite{kazi2019self,kazi2019graph,coates2019mgmc}, where each graph is built from different patient features and encodes a unique latent structure about the population. Patient feature vectors are processed through each graph separately, and fused prior to the decision layer. Multi-graph approaches significantly improve the classification accuracy for CADx, and make GCNs much more robust towards the definition of individual graphs and similarity metrics \cite{kazi2019self,kazi2019graph}. On the other hand, they open up new challenges: i) multiple graphs limit the scalability due to the number of parameters, and ii) the multi-graph fusion layer needs to be carefully designed (e.g. self-attention \cite{kazi2019self} or LSTM-based attention \cite{kazi2019graph}).
An alternative approach to hand-crafting the graph adjacency matrix is to learn it end-to-end, to optimally support the downstream task. Notably, while we focus on CADx in this paper, learning a graph can benefit many other applications as well. In some cases, the learned graph might even be of higher interest than the downstream task, as it can provide important information for interpretability. There is little work so far on graph learning in general \cite{qiao2018graphlearningreview}, and in the field of medicine in particular. Approaches proposed so far, however, are very diverse in nature. For example, Zhan et al. \cite{zhan2019discovery} construct multiple graph Laplacians and optimize the weighting among them during training. Franceschi et al. \cite{franceschi2019icml} sample graph adjacency matrices from a binomial distribution via reparameterization trick and optimize the graph structure via hypergradient descent, demonstrating its efficacy via classification accuracy e.g. in citation networks. Jang et al. \cite{soobeom2019} simultaneously learn an EEG feature representation and a brain connectivity graph based on deterministic graph sampling. Interestingly, they demonstrate that the graph makes sense from a neuroscientific perspective, setting an example that the learned graph can have an intrinsic value regarding model interpretability and knowledge discovery.

\textbf{Contribution:} In this work, we propose a graph convolution-based neural network to perform patient classification, which automatically \textit{learns to predict an underlying patients-graph} that is optimal for the downstream task, e.g. CADx. We show that using a single graph \textit{learned end-to-end} allows both to achieve better performance and reduce the network complexity.
Moreover, our method naturally \textit{applies to inductive settings}, since we learn a function predicting the underlying graph structure which is robust to the introduction of new patients to the current population.
\section{Method}
In this section we describe our proposed model for graph learning and node classification. The advantages of our method are two-fold. First, despite the recent successes of multi-graph methods \cite{kazi2019self,kazi2019graph,coates2019mgmc}, we show that using a graph that has been learned end-to-end allows to achieve a significantly better classification performance. The second advantage is that during graph inference, a single graph is built from multiple features via Euclidean embedding, which drastically reduces the network complexity and solves the scalability problem, as an arbitrary amount of features can be embedded. Compared to CADx GCN models so far \cite{parisot2018disease,vivar2018multi,kazi2019inceptiongcn,kazi2019self}, we thus do not have to restrict the graph's adjacency context to patient meta-feature only (e.g. sex, age), but we can represent and embed a much richer patient representation into the graph structure. In the following paragraphs we detail the architecture of our graph learning module and its use in a classification network.
\begin{figure}[t]
\centering
\includegraphics[width=\textwidth]{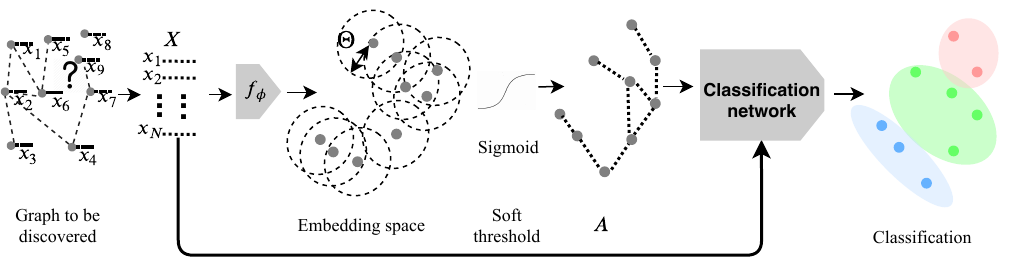}
\caption{Latent-graph learning architecture: Input node features are embedded into a lower dimensional space by a MLP $f_\phi$. The parameter $\Theta$ is a soft-threshold applied to the distances between embedded features in order to build the adjacency matrix $A$. Its outcome provides the population graph model to the GCN, to further learn the node representations for the classification task.}
\label{fig1}
\end{figure}
\subsection{Latent-graph learning}
Given an input set of nodes and associated features $\mathbf{x_{i}} \in \mathbb{R}^d$, the goal is to predict an optimal underlying graph structure which allows employing graph convolutional (GC) layers towards the downstream task, e.g. CADx in this work.\\
Predicting a discrete (e.g. binary) graph structure is a non-differentiable problem. To overcome this limitation and train the graph end-to-end, we represent it as a real-valued, i.e. weighted adjacency matrix $\textbf{A} \in [0,1]^{N\times N}$. With this continuous relaxation we allow the output of any GC layer to be differentiable with respect to the graph structure.

In order to keep the architecture computationally and memory efficient, rather than learning on edge features \cite{soobeom2019,velivckovic2017graph}, we propose to learn a function $\mathbf{\tilde{x_i}} = f_{\phi}(\mathbf{x_i})$ which embeds node input features into a lower dimensional Euclidean space. The edge weight $a_{ij}$ connecting $i^{th}$ and $j^{th}$ nodes is thus directly related to the embedded features distance through the following sigmoid function:
\begin{equation}
    a_{ij}=\frac{1}{1+e^{-t \left( \|\mathbf{\tilde{x}_{i}}-\mathbf{\tilde{x}_{j}}\|_2 + \theta \right)} }
\label{eq:sigmoid}
\end{equation}
where $\theta$ is a threshold parameter and $t$ is a temperature parameter pushing values of $a_{ij}$ to either $0$ or $1$. Values of $t$ and $\theta$ are both optimized during training. We use a simple Multilayer Perceptron (MLP) as our function $f_{\phi}$. 
In Figure \ref{fig1} we illustrate the graph learning pipeline.
\subsection{Classification model}
We use our Latent-Graph learning to perform node classification. Given a set of $N$ patients and associated set of multi-modal features $ \textbf{X} \in \mathbb{R}^{N \times d_{1}} $, the task is to predict the corresponding labels $y_i, i=1\dots N$. 
To this end, we build a classification model composed by few graph convolutional layers followed by a fully connected layer to predict the patient label. To build the graph $\textbf{A}$  with our latent-graph learning module, our model simply requires the full patient feature vectors $X$. In particular, we make use of the spatial GC layer defined as:  

\begin{equation}
    \mathbf{H_{l+1}} = \sigma( \mathbf{D}^{-1}\mathbf{A}\mathbf{H}_{l}\mathbf{W})
\end{equation} 
where $\sigma$ is a non-linear activation function, $\textbf{D}$ is a normalization matrix with $d_{ii} = \sum_{ij} a_{ij}$, $\mathbf{H}_{l}$ is the output activation of the previous layer, and $W$ are the model filters to be learnt. As mentioned before, both the latent-graph learning and classification MLPs are trained in an end to end manner. 
\begin{figure}[t!]
\centering
\includegraphics[width=0.355\textwidth]{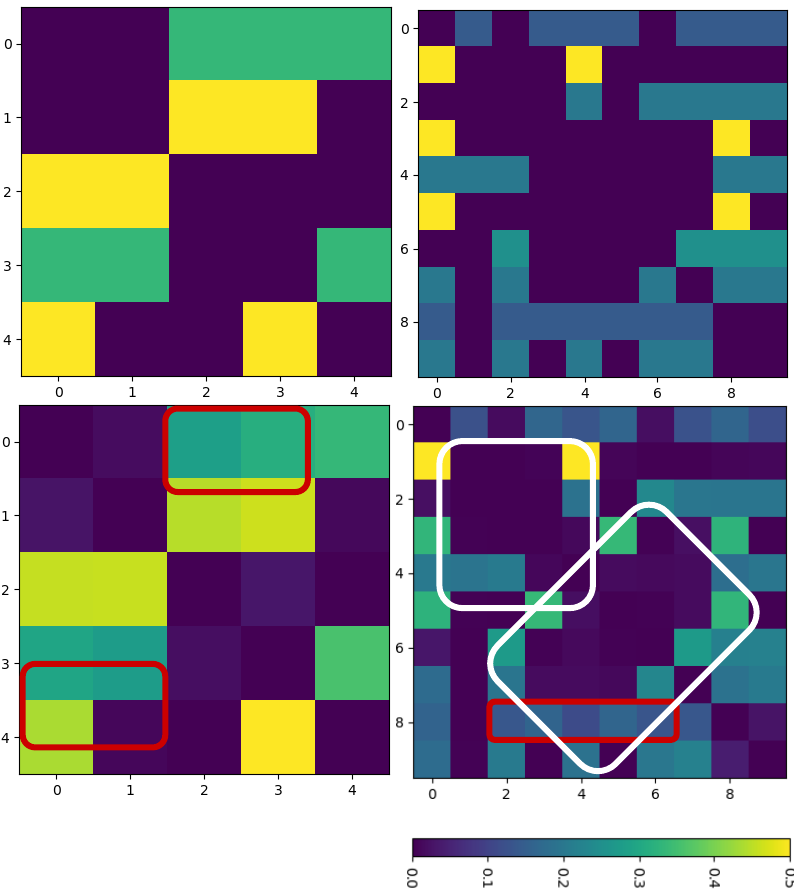}
\includegraphics[width=0.575\textwidth]{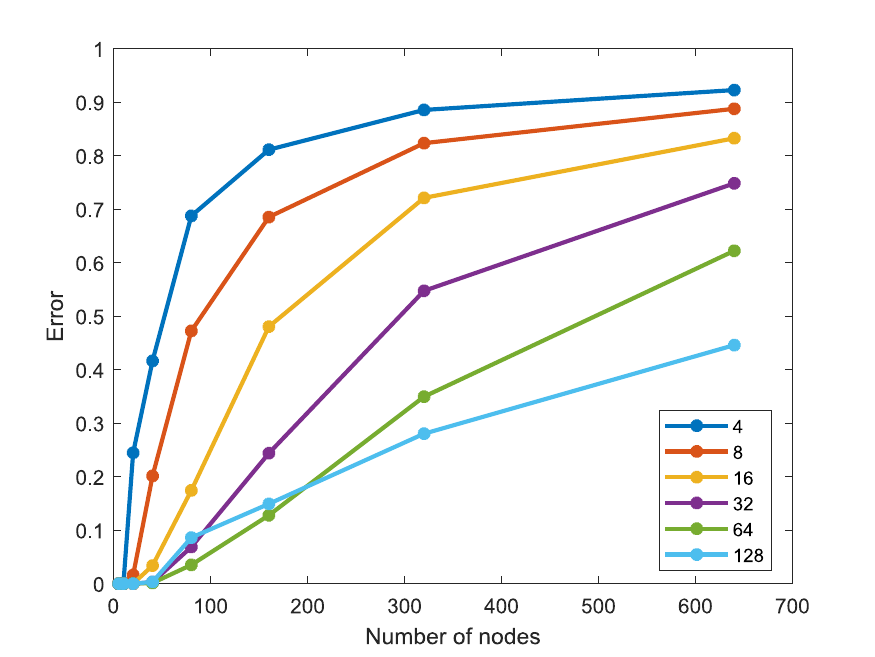}
\caption{\textbf{Left:} the ground truth graphs (top) and output of the graph optimization (bottom) for 5 and 10 nodes setting. The red boxes show errors made by our method, and white boxes show successful adjacency reconstruction. \textbf{Right:} the mean squared error between predicted and ground truth vectors as a function of number of nodes. Each curve represents different embedding dimensions.}
\label{fig:simulated}
\end{figure}
% \\
%
\section{Experiments and Results}
We start this section with a proof of concept example showing the ability of our method to retrieve the ground truth graph. The remaining of the section is dedicated to comparison with the-state-of-the-art methods on two publicly available medical datasets.

\subsection{Proof of concept:}
In order to test the graph learning part of our model, we design an experimental setup on simulated data.
We define a simple optimization problem in which the task is to regress a target vector $y_i$ associated to each node.\\
\textbf{Setting.}
Consider to be given a randomly generated graph $\mathbf{\mathcal{G}}$ of $N$ nodes with associated features $\textbf{X}$. 
We define the ground-truth vector of the $i^{th}$ node as the sum of its neighbors' features, $y_i = \sum_{j \in \mathcal{N}_i}x_{ij}$.
Further, we initialize the feature matrix as $\textbf{X}= \textbf{I} \in \mathbb{R}^{N \times N}$. Since our focus is to analyze the graph learning part, identity features are simple and orthogonal, which results in a minimum impact of the feature space on the task. As a result, only one possible output graph structure can minimise the categorical cross-entropy loss to zero.\\
\textbf{Task.} 
With this setting, we formalize the following optimization problem:
\begin{equation}
    \arg \min_\Phi \sum_i (\mathbf{y}_i - \textbf{A}(\Phi,\textbf{X}) \mathbf{X}_{:i})^2
    \label{eq:opt}
\end{equation}
where $\textbf{A}(\Phi,\textbf{X})$ is our latent-graph learning module with input node features $\textbf{X}$ and optimized parameters $\Phi$. The main challenge in optimizing eq. \ref{eq:opt} is to choose the neighborhood between the nodes.\\
\textbf{Results.} In figure \ref{fig:simulated} we show results of the optimization. We analyze the behaviour of our method at increasing number of nodes. From the right plot we can see how the embedding size plays an important role on the ability of the method to retrieve the underlying graph. On the left we can appreciate how our latent-graph learning module is able to retrieve the relevant (ground-truth) affinity graph on two simulated toy graphs with 5 and 10 nodes.

\subsection{Comparisons}

After our proof of concept on simulated data, we now show results of the proposed method on two publicly available medical datasets. We benchmark our classification results against three baselines and three comparative methods, which will be described in the following.
We choose TADPOLE and UKBB mainly due to the difference is their size, which helps us analyze the adaptability of our proposed graph learning technique to smaller and larger datasets in medicine. \\
\textbf{TADPOLE:} This dataset \cite{marinescu2018tadpole} is a subset of the Alzheimer's Disease Neuroimaging Initiative (adni.loni.usc.edu), comprising 564 patients with 354-dimensional multi-modal features. The task is to classify each patient into Cognitively Normal (CN), Mild Cognitive Impairment (MCI), or Alzheimer's Disease (AD). Features are extracted from MR and PET imaging, cognitive tests, cerebro-spinal fluid (CSF) and clinical examinations. \\
\textbf{UKBB:} UK Biobank data provides pre-computed structural, volumetric and functional features of the brain, which are extracted from MRI and fMRI images using Freesurfer \cite{fischl2012freesurfer}. Here, we task the network to predict the age for each patient. We use a subsample of UKBB data \cite{miller2016multimodal}, consisting of 14,503 patients with 440 features per individual. We quantize the patients' age into four decades from age 50-90 as classification targets.

\textbf{Error metrics.} We use two standard error metrics to evaluate the performance of the network; accuracy and area under the ROC-curve (AUC). In multi-class classification tasks AUC is computed as the average of the AUC of each class versus all the others.

\begin{table}[t!]
\centering
\setlength{\tabcolsep}{4pt}
\caption{Comparison with some baseline methods on classification task on TADPOLE and UKBB datasets.}
\label{tab:baselines}
\begin{tabular}{lcc|cc}
\hline
&\multicolumn{2}{c}{TADPOLE}&\multicolumn{2}{|c}{UKBB}\\
\hline
Baselines & Accuracy & AUC & Accuracy & AUC \\
\hline
Linear classifier&  70,22 $\pm$ 06,32 & 80.26 $\pm$ 04.81 & 59.66$\pm$ 1.170 & 80.26 $\pm$ 00.91\\
Spectral-GCN \cite{parisot2018disease} &81.00 $\pm$ 06.40&74.70 $\pm$ 04.32  &OOM&OOM\\
DGCNN\cite{wang2019dynamic} &  84.59 $\pm$ 04.33 & 83.56 $\pm$ 04.11 & 58.35 $\pm$ 00.91& 76.82 $\pm$ 03.03\\
Proposed& \textbf{92.91 $\pm$ 02.50}&94.49 $\pm$ 03.70 &64.35 $\pm$ 01.11 & 82.35 $\pm$ 00.36\\
\hline
\end{tabular}
\end{table}

\textbf{Baselines.} Table \ref{tab:baselines} presents the results of some baselines in comparison with our proposed method. As a linear and non-graph baseline technique, we use a ridge regression classifier. Spectral-GCN \cite{parisot2018disease} performs graph based convolution in the Fourier domain, but requires a pre-defined graph. Dynamic graph CNN (DGCNN) \cite{wang2019dynamic} constructs a kNN graph on the output activation during training, however the graph is not learned. All the graph-based techniques perform better than the ridge classifier (\ref{tab:baselines}). It can be further observed that a pre-defined graph may indeed be a sub-optimal choice, since Spectral-GCN performs worse than both DGCNN and the proposed method. Importantly, due to the polynomial filter approximation of the graph Laplacian, Spectral-GCNs can have too high memory demands, which is why UKBB could not be evaluated with this model. Further, our proposed model outperforms all the baselines with a margin of 8.32\% and 6\% for TADPOLE and UKBB respectively.

\begin{table}[b!]
\centering
\caption{Comparison with state-of-the-art graph convolutional based methods on TADPOLE and UKBB datasets. P denotes the number of model parameters.}

\setlength{\tabcolsep}{3pt}
\label{tab:comparative}
\begin{tabular}{lccc|ccc}
\hline
&\multicolumn{3}{c}{TADPOLE}&\multicolumn{3}{|c}{UKBB}\\
\hline
Method & Accuracy & AUC & P&Accuracy& AUC& P \\
\hline
Multi-GCN \cite{kazi2019self} &76.06$\pm$0.72&90.32$\pm$4.85&46k&OOM&OOM&46k\\
InceptionGCN\cite{kazi2019inceptiongcn}&84.11$\pm$4.50&88.39$\pm$4.16&58k&OOM&OOM&58k\\
DGM\cite{kazi2020differentiable}&91.05$\pm$5.93&\textbf{96.86$\pm$1.81}&6k&61.59$\pm$ 1.05&79.32$\pm$0.95&14k\\
Proposed &\textbf{92.91$\pm$2.50}&94.49$\pm$3.70& \textbf{2k}&\textbf{64.351$\pm$1.11}&\textbf{82.352$\pm$0.37}&\textbf{11k}\\
\hline
\end{tabular}
\end{table}

\textbf{Comparative methods.} In table \ref{tab:comparative} we show the performance of our model with respect to some state-of-the-art methods on the considered datasets. We choose Multi-GCN \cite{kazi2019self}, which uses multiple graphs from different set of features, and reportedly performed very well on datasets similar to TADPOLE. InceptionGCN \cite{kazi2019inceptiongcn} is a state of the art GCN method on TADPOLE. Differentiable Graph Module (DGM) \cite{kazi2020differentiable} is a recently proposed graph learning method. As can be seen in table \ref{tab:comparative}, the proposed model outperforms all comparative methods. For the  much larger UKBB dataset, we again received out of memory (OOM) errors for the spectral GCN methods Multi-GNC and InceptionGCN. 
We further report the number of parameters required by each compared model in table \ref{tab:comparative}.
All the experiments are performed with ten-fold stratified cross-validation. The low standard deviation of our proposed method further shows the improved robustness of our model.

\textbf{Out of sample extension.} 
Our method does not directly optimize a graph for a given population but it rather learn a $f_{\phi}$ that predicts the graph from input patients features. As such, it is easily extendable to previously unseen test patients, to enable inductive inference. Unseen patients can thus be added during testing time, and will be embedded into the lower dimensional space for graph representation. 
For a comparison to state of the art, we select DGCNN and DGM which are both inductive graph methods as well. Table \ref{tab:inductive} shows the accuracy of classification, given a data split of 90\% training data vs 10\% testing data. The higher accuracy and lowest standard deviation for our method confirms the superiority, robustness and precision of our model in a fully inductive setting.

\begin{table}[b!]
    \centering
    \caption{Classification accuracy score for out of sample extension in TADPOLE and UKBB datasets.}
    \setlength{\tabcolsep}{6pt}
    \begin{tabular}{lccc}
    \hline
        Method&DGCNN&DGM&Proposed\\%& ABIDE\\
         \hline
         TADPOLE&82.99 $\pm$ 04.91&88.12 $\pm$ 03.65& \textbf{91.85 $\pm$ 02.62}\\
         UKBB&51.84 $\pm$ 08.16& 53.37 $\pm$ 07.94& \textbf{63.91  $\pm$ 01.49}\\
        \hline  
    \end{tabular}

    \label{tab:inductive}
\end{table}
\begin{figure}[t!]
\centering
\includegraphics[width=0.8\textwidth]{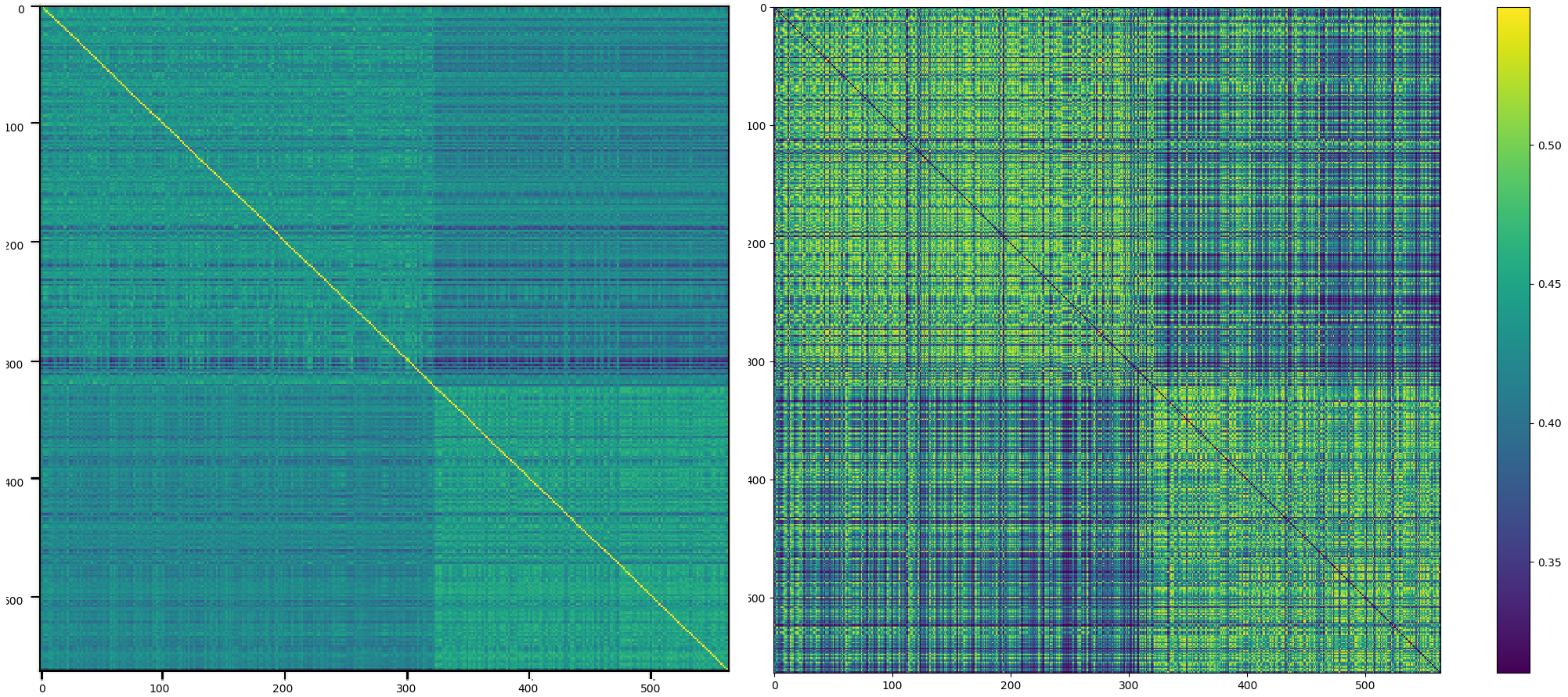}
\caption{The figure shows the ground truth graph (left) and learnt graph on the (right). The graph is shown for all the point in the dataset for random fold. The colorbar shows the affinity of each point with other.}
\label{fig:qualitative}
\end{figure}

\textbf{Qualitative results.}
Our quantitative results show the superiority of our proposed method, which learns a graph along with the task in an end to end fashion. However, from a qualitative point of view, it is challenging to evaluate the learnt graph, because it is not possible to compute a groundtruth graph. As an approximation, we generate a reference graph for TADPOLE from a weighted summation of the individual graphs from different modalities as is described in Multi-GCN \cite{kazi2019self}. Figure \ref{fig:qualitative} shows the ground truth graph on the left and the learned graph. Interestingly, even though both the graphs are not completely identical, the overall structure shows similarities with one another. Remaining differences in the graph structure are expected, as these probably explain the improved classification performance to a degree. However, a proper interpretation from a medical perspective would require an assessment by experts, which we suggest as future work.

\textbf{Implementation details.}
For both datsets we use the same architecture composed by two convolution layers ($16 \rightarrow 8$) and a final MLP ($32 \rightarrow 16 \rightarrow $ \#classes) for obtaining the classification score. We use RElU non linearity and dropout (keep rate 0.9) after each layer except the last. Following \cite{parisot2017spectral} we reduce the input feature dimensions via \textit{recursive feature elimination} \cite{granitto2006recursive} to 30 and 200 for TADPOLE and UKBB respectively and apply standard normalization.
We use the Adam optimizer \cite{kingma2014adam} to minimize the categorical cross-entropy loss with a learning rate of 0.01 reduced to 0.0001 at the intervals of 100 epochs in a piecewise constant fashion. The number of epochs=600. All the experiments are implemented in TensorFlow and performed using a commercial workstation and GPU (Titan Xp, 12GB VRAM).\\

\section{Discussion and Conclusion}
In this paper, we propose a model capable of learning the optimal relationships between the patients towards the downstream task, e.g. CADx and age prediction in medicine in this work. The entire model is trained in an end-to-end fashion, backpropagating directly through the graph adjacency. This is achieved using a soft thresholding technique with the learnable threshold $\Theta$ and the temperature parameter. Such a setting allows the update of each edge weight with respect to the loss.

As a proof of concept, we showed experiments on the simulated data where we successfully learned the given ground truth graph. Further, we showed applications to disease prediction on TADPOLE data and age prediction on UK Biobank data. In all the experiments our model performed better than baseline and state of the art approaches. Our model generalizes to inductive setting and outperforms other state of the art methods. At the same time, our model only contains up to two orders of magnitude less parameters than state of the art single- and multi-graph methods. It can be concluded from all of our experiments that a pre-defined graph might not be optimal, neither in a single- nor in a multi-graph setting. Instead, an end-to-end learning of the graph adjacency can lead to significant benefits in downstream tasks like classification. Further, the graph structure itself may have an intrinsic value, e.g. for better interpretability and knowledge discovery in the medical dataset.

In the proposed method we learned a global threshold for the entire population, but this may not be necessarily optimal. A single threshold operating globally on the Euclidean embedding might neglect the heterogeneity of the embedding structure. Therefore, learning a dedicated and patient-specific neighborhood threshold for each node might be the next step. Likewise, the Euclidean space embedding may not be optimal to learn a semantically meaningful graph either. Recent works have shown that the appropriate isometric space for embedding graphs is a negatively curved, i.e. a hyperbolic space \cite{chamberlain2017neural}. Another direction to explore is the interpret the graph as well as the model.

In summary, we have proposed a novel graph learning method which makes it unnecessary to pre-define graph adjacencies. Until now, this was a prerequisite for employment of GCN models in medicine so far, but the definition of patient similarity metrics was not often well-motivated. The proposed graph learning method thus paves the way for further employment of graph methods in clinical decision support systems, which consistently demonstrate higher classification performances than the current state of the art in medicine.
\section{Acknowledgement}
The study was carried out with financial support of TUM-ICL incentive funding, Freunde und F{\"o}rderer der Augenklinik, M{\"u}nchen, Germany and ERC Consolidator grant No. 724228 (LEMAN) and German Federal Ministry of Education
and Health (BMBF) in connection with the foundation of the German Center for Vertigo and Balance Disorders (DSGZ) [grant number 01 EO0901]. The UK Biobank data is used under the application id 51541.\\

\bibliographystyle{splncs04}
\bibliography{main}

\end{document}